\documentclass{article}
\usepackage{spconf,amsmath,graphicx}
\usepackage{amssymb}
\usepackage{amssymb}
\usepackage{multirow}
\usepackage{tabularx}
\usepackage{enumitem}
\setlist{nosep, leftmargin=14pt}
\usepackage{pifont}

\usepackage{mwe} 

\title{Reference-based Magnetic Resonance Image Reconstruction Using Texture Transformer}
\twoauthors
  {Pengfei Guo}
	{Department of Computer Science\\ Johns Hopkins University\\ Baltimore, MD}
  {Vishal M. Patel}
	{Department of Electrical and Computer Engineering\\ Johns Hopkins University\\ Baltimore, MD}\usepackage{amsmath}
\begin{document}
\ninept
\maketitle
\begin{abstract}
Deep Learning (DL) based methods for magnetic resonance (MR) image reconstruction have been shown to produce superior performance in recent years. However, these methods either only leverage under-sampled data or require a paired fully-sampled auxiliary modality to perform multi-modal reconstruction. Consequently, existing approaches neglect to explore attention mechanisms that can transfer textures from reference fully-sampled data to under-sampled data within a single modality, which limits these approaches in challenging cases. In this paper, we propose a novel Texture Transformer Module (TTM) for accelerated MRI reconstruction, in which we formulate the under-sampled data and reference data as queries and keys in a transformer. The TTM facilitates joint feature learning across under-sampled and reference data, so the feature correspondences can be discovered by attention and accurate texture features can be leveraged during reconstruction. Notably, the proposed TTM can be stacked on prior MRI reconstruction approaches to further improve their performance. Extensive experiments show that TTM can significantly improve the performance of several popular DL-based MRI reconstruction methods.
\end{abstract}
\section{Introduction}
\label{sec:intro}
Magnetic resonance imaging (MRI) is one of the most widely used noninvasive medical imaging techniques. Different pulse sequences provide various contrast mechanisms for visualizing anatomical structures and tissue functions. However, due to the acquisition hardware limitation, relatively slow data acquisition process of MRI impedes its development in many clinical applications. In spite of the
exploitation of advanced hardware and parallel imaging, a conventional approach that can shorten the image acquisition time is leveraging Compressed Sensing (CS)~\cite{cs} to reconstruct MR images from under-sampled $k$-space data. Unfortunately, though CS reconstruction algorithms are able to recover images, they lack the ability of recovering noise-like textures and introduce high-frequency oscillatory artifacts when large errors are not reduced during optimization~\cite{cs-artifacts}.
\begin{figure}[t]
	\centering
	\includegraphics[width=0.75\columnwidth]{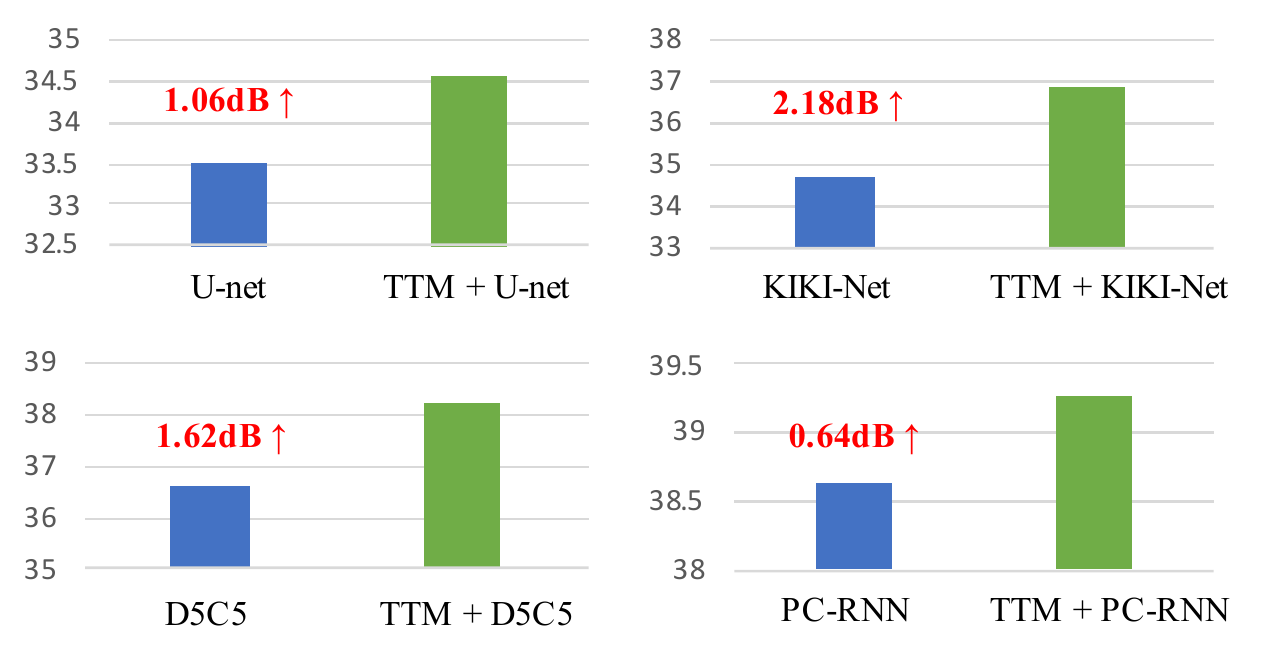}
	\vskip-10pt	
	\caption{The performance changes of the popular MRI reconstruction models after combing them with the  proposed TTM in term of PSNR. Here, the acceleration factor (AF) was set to 4.} \label{fig1}
	\vskip-20pt
\end{figure}

Recent emergence of DL-based methods in addressing this inverse problem of MR image reconstruction has significantly improved the quality of recovered images in an efficient manner. The research on MRI reconstruction has mainly conducted on two
paradigms, including single image reconstruction (SIR), and guidance-based image reconstruction (GIR). The SIR problem aims to recover  fully-sampled images from a single under-sampled $k$-space data. Several SIR methods~\cite{unet,kiki,d5c5,pcrnn,oucr} have been proposed and have demonstrated superior performance over the CS-based methods. However, in challenging cases, SIR methods often result in blurry artifacts, since the anatomic textures have been excessively degraded in the $k$-space under-sampling process. Several methods have focused on guidance-based image reconstruction, which explores textures from a given reference fully-sampled auxiliary modality to produce visually pleasing results~\cite{gir2,gir3,gir4,gir5}. However, those approaches usually require a multi-modal dataset for training. For example, reconstructing a $T_2$-weighted image usually requires a paired $T_1$-weighted image as the reference auxiliary modality, which significantly increases the difficulty of data acquisition and impedes the practicality of GIR methods. 

To address these issues and inspired by the successful applications of transformers on image restoration tasks~\cite{ttsr, Attention}, we propose a novel Texture Transformer Module (TTM) for MR image reconstruction. Specifically, we formulate the
extracted features from the under-sampled and reference image as the query
and key in a transformer. The produced attention maps are used to transfer the features from the reference image into the features extracted from the under-sampled data. To the best of our knowledge, the proposed TTM is the the first attempt to explore relevance and transfer texture features between the reference image and the under-sample input within a single MRI modality using a transformer. As shown in Fig.~\ref{fig1}, the design of TTM provides an accurate way to search and transfer relevant textures from reference to under-sampled data and combining the TTM with a reconstruction backbone can significantly improve the quality of recovered images.

\section{Methodology}
\label{sec:meth}

\begin{figure}[t]
	\centering
	\includegraphics[width=\columnwidth]{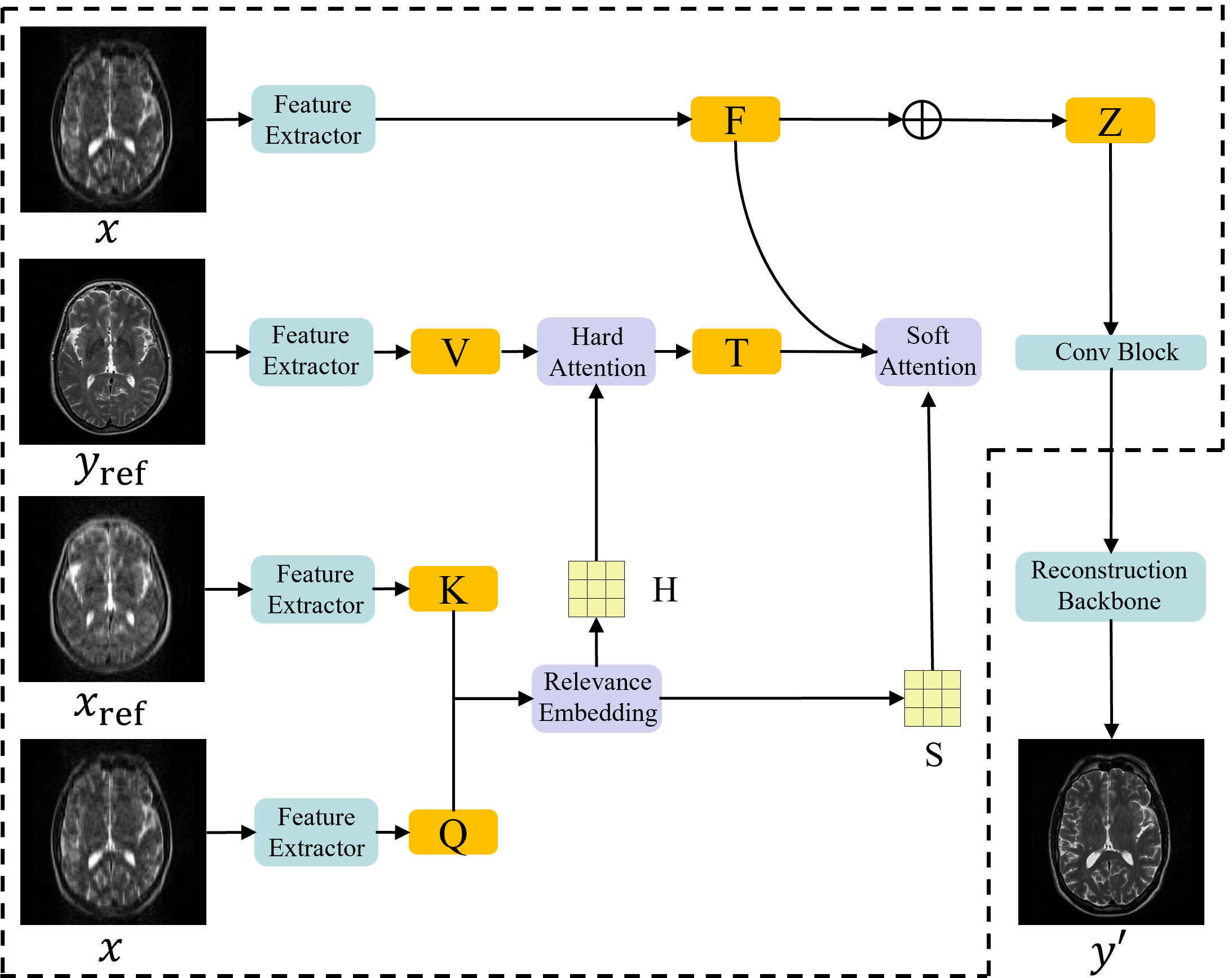}
	\vskip-10pt	
	\caption{An overview of the proposed texture transformer module. $Q, K$ and $V$ are the
		texture features extracted from input under-sampled image $x$, the under-sampled reference image $x_{\text{ref}}$, and the fully-sampled reference image $y_{\text{ref}}$, respectively.  $H$ and $S$ indicate the hard and soft attention map, respectively.  $F$ is the under-sampled features extracted from a $x$ and is further fused with the transferred texture features $T$ to generate the synthesized features $Z$. Finally, a synthesized under-sampled image $x^\prime$ is generated from $Z$ via a convolutional block.} \label{fig2}
\end{figure}
\subsection{MR image Reconstruction}
Accelerated magnetic resonance image reconstruction is an inverse problem in which the objective is to reconstruct a fully-sampled image from under-sampled $k$-space data. The data acquisition process can be formulated as follows:
\setlength{\belowdisplayskip}{0pt} \setlength{\belowdisplayshortskip}{0pt}
\setlength{\abovedisplayskip}{0pt} \setlength{\abovedisplayshortskip}{0pt}
\begin{equation} \label{eq:1}
\begin{aligned}
&x = F^{-1}(F_{D}y + \epsilon),
\end{aligned}
\end{equation}
where $x$ denotes the observed under-sampled image, $y$ is the fully-sampled image, and $\epsilon$ denotes noise. We denote the Fourier transform matrix and its inverse as $F$ and $F^{-1}$, respectively. $F_D$ represents the under-sampling Fourier encoding matrix that is defined as the multiplication of the Fourier transform matrix $F$ with a binary undersampling mask matrix $D$. The acceleration factor (AF) controls the ratio of the amount of $k$-space data required for a fully-sampled image to the amount collected in an accelerated acquisition. The goal is to estimate $y$ from the observed under-sampled image $x$.

\subsection{Texture Transformer Module}
Fig.~\ref{fig2} gives an overview of the proposed texture transformer module. $x$, $x_{\text{ref}}$, and $y_{\text{ref}}$ represent the input under-sampled image, the under-sampled reference image, and the fully-sampled reference image, respectively. Since we apply the same under-sampling operations, $x_{\text{ref}}$ is domain-consistent with $x$. Inspired by~\cite{ttsr}, the proposed TTM consists of five parts as follow: feature extractor, relevance embedding module, hard attention module for feature transfer, soft attention module for feature synthesis, and a convolutional block for producing output. TTM takes $x$, $x_{\text{ref}}$, and $y_{\text{ref}}$ as input, and outputs a synthesized under-sampled image $x^\prime$, which contains real and imaginary channels and can be further used to generate the fully-sampled prediction $y^\prime$ by the reconstruction backbone. In what follows, we describe different parts of the proposed method in detail.

\noindent {\bf{Feature Extractor. }} Extracting accurate and proper texture information plays an essential role to facilitate the downstream fully-sampled MR image reconstruction. Rather than using a pre-trained classification model, to encourage the joint feature learning, we introduce a learnable feature extractor and its parameters are updated during the training. Query $(Q)$, Key $(K)$, and Value $(V)$ are three basic elements of the attention mechanism in a transformer, and can be formulated as follows: 
\begin{equation} \label{eq:2}
\begin{aligned}
Q=FE(x), K=FE(x_{\text{ref}}), V=FE(y_{\text{ref}}),
\end{aligned}
\end{equation}
where $FE(\cdot)$ denotes the output of feature extractor. The proposed feature extractor consists of 4 convolutional blocks. Each convolutional block contains a 2D convolutional layer with kernel size as 3, stride as 1, and padding as 1 followed by a Rectified Linear Unit (ReLU) activation layer. To enable high flexibility at expressing the structure of the data, the channel dimension of the output of feature extractor is increased to 64.

\noindent {\bf{Relevance Embedding Module. }} To embed the relevance between the $x$ and $x_{\text{ref}}$, the similarity between $Q$ and $K$ is estimated by the relevance embedding module. $Q$
and $K$ are unfolded into patches with size of $16 \times 16$ and we denoted as $q_i$ and $k_j$, where $i$ and $j$ represent the spatial locations of patches. For each patch pair between $Q$ and $K$, we can calculate the relevance score $r_{i,j}$ by the normalized inner product as follows:
\begin{equation} \label{eq:3}
\begin{aligned}
r_{i,j} = \left\langle \frac{q_i}{\|q_i\|}, \frac{k_j}{\|k_j\|} \right\rangle.
\end{aligned}
\end{equation}
Obtained relevance scores are further used to generate hard and soft attention
maps.

\begin{figure*}[t]
	\centering
	\includegraphics[width=.93\textwidth]{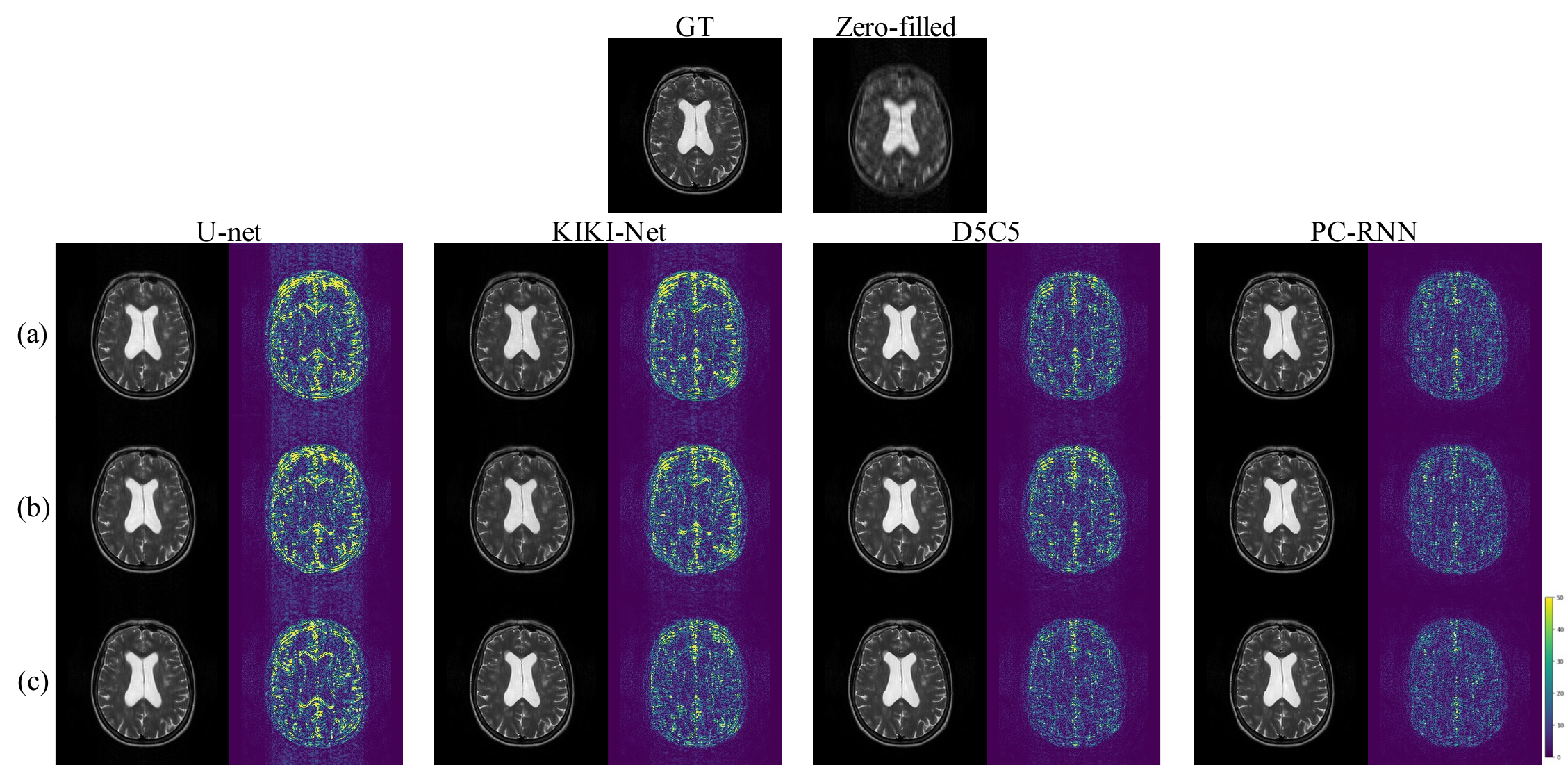}
	\vskip-10pt	
	\caption{Visual comparison of different MRI reconstruction methods on IXI~\cite{IXI} testing set among 3 strategies --- (a) Original, (b) Ref+Original, (c) TTM+Original. The second column of each sub-figure shows the absolute image difference between reconstructed images and the ground truth.} \label{fig3}
\end{figure*}
\begin{table*}[t!]
	\centering
	\setlength{\tabcolsep}{11pt}
	\caption{PSNR/SSIM comparison among different strategies on the IXI dataset. AVG. Gain represent the performance changes compared to the Original.}\label{tab1}
	\begin{tabular}{c|c|c|c|c|c}
		\hline
		\multirow{2}{*}{Strategy} & \multicolumn{4}{c|}{Reconstruction Backbone }                     & \multirow{2}{*}{AVG. Gain} \\ \cline{2-5}
		& U-net~\cite{unet}          & KIKI-Net~\cite{kiki}       & D5C5~\cite{d5c5}           & PC-RNN~\cite{pcrnn}          &                            \\ \hline
		Original                  & 33.49 / 0.9131 & 34.73 / 0.9456 & 36.61 / 0.9656 & 38.64 / 0.9797 & -                          \\ \hline
		Ref+Original            & 33.52 / 0.9134 & 34.70 / 0.9438 & 36.38 / 0.9624 & 38.61 / 0.9796 & $-$ 0.07 $\downarrow$ / $-$ 0.0012 $\downarrow$           \\ \hline
		TTM+Orignal (Our)       & 34.25 / 0.9238 & 36.91 / 0.9682 & 38.23 / 0.9744 & 39.28 / 0.9818 & $+$ 1.30 $\uparrow$ / $+$ 0.0111 $\uparrow$              \\ \hline
	\end{tabular}
\end{table*}

\noindent {\bf{Hard Attention Module. }} Hard attention module is designed to
search the relevant fully-sampled texture features $V$ from the $y_{\text{ref}}$. Rather than taking a weighted sum of $V$ for each query $q_i$, only the most relevant feature in $V$ is transferred for each query $q_i$. Such design can prevent blurry outputs and increase the ability of transferring full-sampled texture features~\cite{ttsr}. Specifically, the $i$-th element $h_i$ of the hard-attention map $H$ can be calculated from the relevance score as follows:
\begin{equation} \label{eq:4}
\begin{aligned}
h_{i} = \underset{j}{\mathrm{argmax}}\: r_{i,j}.
\end{aligned}
\end{equation}
Each $h_i$ can be treated as a hard index and indicates the most relevant position in the  $y_{\text{ref}}$ for the $i$-th position in $x$. Subsequently, $V$ is also unfolded into patches with size of $16 \times 16$ and the transferred texture features $T$ from $y_{\text{ref}}$ can be selected by using the hard-attention map as the indices. We denote $t_i$ as the patch of $T$ in the $i$-th position. Then, $t_i$ is selected from the $h_i$-th position of $V$ and can be expressed as $t_{i} = V_{h_i}$.

\noindent {\bf{Soft Attention Module. }} Soft attention module aims to generate the synthesized features $Z$ from the transferred texture features $T$ and under-sampled features $F$. To be wary of the confidence of the transferred texture features, during synthesis process, a soft attention map $S$ is calculated after relevance embedding as follow:
\begin{equation} \label{eq:5}
\begin{aligned}
s_{i} = \underset{j}{\max}\: r_{i,j},
\end{aligned}
\end{equation}
where $s_i$ is the $i$-th position of the soft-attention map $S$. To utilize more information from the input image,  transferred texture features $T$ and under-sampled features $F$ are first fused, then the fused features are element-wise multiplied by the soft attention map $S$ and added back to $F$ to produce the synthesized features $Z$. This synthesis process can be expressed as follows:
\begin{equation} \label{eq:6}
\begin{aligned}
Z = F + \text{Conv} (F\oplus T) \otimes S,
\end{aligned}
\end{equation}
where Conv represents a convolutional layer. $\oplus$ and $\otimes$ denote the element-wise multiplication and channel-wise concatenation between feature maps. Finally, the channel dimension of the synthesized features $Z$ is reduced to 2 (e.g. real and imaginary channels) by a Conv Block. The synthesized under-sampled image $x^\prime$ that contains transferred texture feature from fully-sampled reference image $y_{\text{ref}}$ can be further used as input of other DL-based MR image reconstruction backbones and boosts the performance of the downstream reconstruction process.

\section{Experiments and Results}
\subsection{Evaluation and Implementation Details}
\noindent {\bf{Dataset. }} Experiments are conducted on the IXI~\cite{IXI} dataset. $T_2$-weighted MRI data from 135 subjects are analyzed, where 100 subjects’ data
are used for training, 10 subjects’ data are used for validation, 10 subjects’ data are used as reference and the remaining 15 subjects’ data are used for testing. For each subject, there are approximately 130 axial cross-sectional images that contain brain tissues for $T_2$-weighted MR sequence. We use mutual information as the metric to create input-reference matches. 

\noindent {\bf{Implementation Details. }} All models were trained using the $\ell_{1}$ loss with Adam optimizer based on the following hyperparameters: initial learning rate of $1.5 \times 10^{-4}$ then reduced by a factor of 0.9 every 5 epochs; 50 maximum epochs; batch size of 8; number of RNN iteration of 5 (for PC-RNN~\cite{pcrnn} only).  Hyperparameter selection is performed on the IXI~\cite{IXI} validation dataset. The sampling mask function with 4$\times$ acceleration is used for simulating the $k$-space measurements. Structural similarity index measure (SSIM) and peak-signal-to-noise ratio (PSNR) are used as the evaluation metrics for comparisons.

\subsection{MR Image Reconstruction Results}
To evaluate the effectiveness of the proposed TTM, we observe the performance change of adding TTM to four popular MRI reconstruction methods, including U-net~\cite{unet}, KIKI-Net~\cite{kiki}, D5C5~\cite{d5c5}, and PC-RNN~\cite{pcrnn}. For a fair comparison, U-net~\cite{unet} is modified for data with real and imaginary channels and a data consistency (DC) layer is added at the end of the network. KIKI-Net~\cite{kiki} conducts interleaved convolution operation on the image and the $k$-space, which demonstrates the benefit of cross-domain reconstruction. D5C5~\cite{d5c5}, which consists of several cascaded CNNs, is considered as one of the most popular DL-based MRI reconstruction methods. PC-RNN~\cite{pcrnn} learns the mapping in an iterative way by convolutional recurrent neural networks (CRNN) from three different scales, which has achieved state-of-the-art performance on both PSNR and SSIM in recent years.

Table~\ref{tab1} shows the results corresponding to four different reconstruction methods evaluated on the IXI dataset. We first compare the performance of four MRI reconstruction methods without any modifications and denote them as \textbf{Original} in Table~\ref{tab1}. It is also possible
to naively leverage information provided by the reference images. In this case, we just concatenate the reference image with the input as additional channels. The results corresponding to this strategy are shown in Table~\ref{tab1} under the label \textbf{Ref+Original}. Finally, we can obtain models that use the proposed TTM to transfer texture information and denote experiments as \textbf{TTM+Original}.

\begin{figure}[h]
	\centering
	\includegraphics[width=\columnwidth]{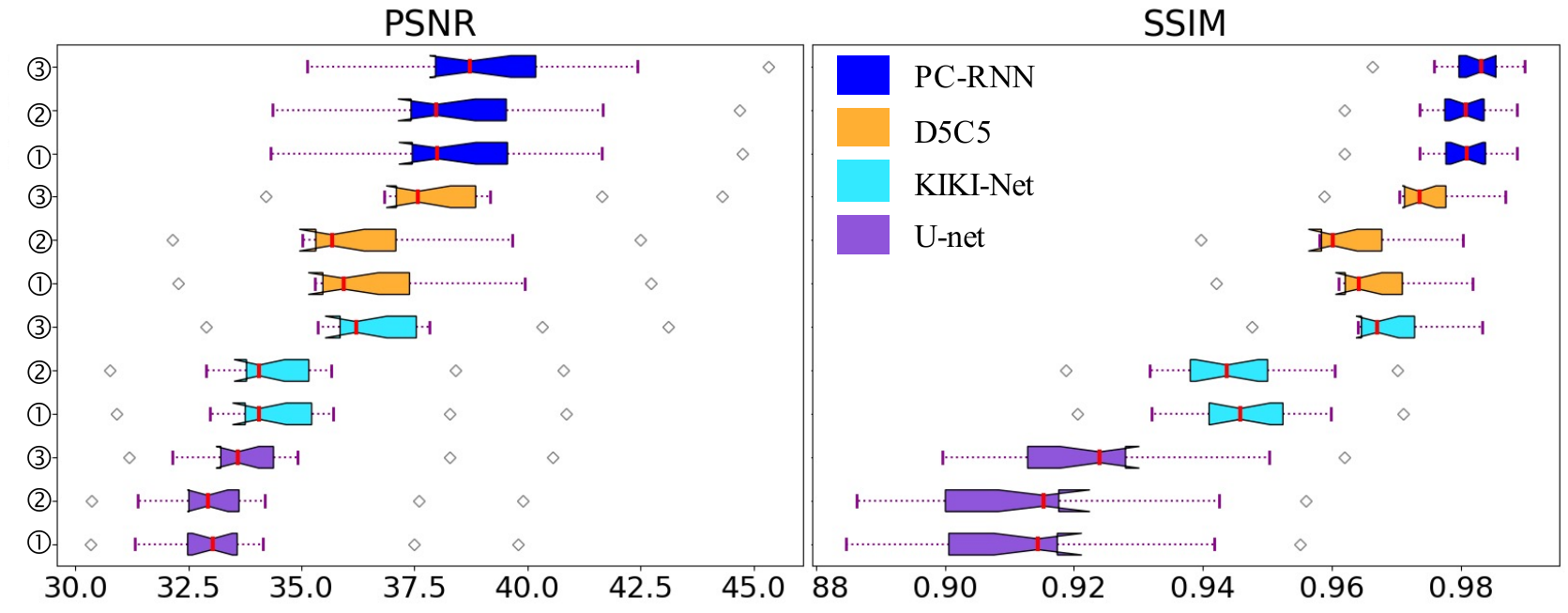}
	\vskip-10pt	
	\caption{Boxplot of reconstruction performance of different strategies among four methods. \ding{172}, \ding{173}, and \ding{174} represent Original, Ref+Original, and TTM+Original, respectively.} \label{fig4}
	\vskip-15pt
\end{figure}

From Table~\ref{tab1}, we can make the following observations: (i) Naively treating the reference image as an additional input cannot improve the reconstruction performance. Due to lacking an efficient mechanism of searching and transferring the texture information in reference images, it results in slightly lower performance across four different methods. (ii) By using the proposed TTM, we are able to achieve obvious improvements as compared to models that naively use the reference images and the original models. (iii) The gain in performance  achieved by TTM are consistent across different reconstruction methods, as shown in Fig.~\ref{fig4}. Even for the state-of-the-art CRNN method (i.e. PC-RNN~\cite{pcrnn}), the proposed TTM still can improve the reconstruction quality by 0.64 dB in term of PSNR. Fig.~\ref{fig3} shows the
qualitative performance of different methods on the $T_2$-weighted images from the IXI~\cite{IXI} dataset. It can be observed that the models that make use of the proposed TTM yield reconstructed images with remarkable visual similarity to the GT images compared to the other alternatives (see sub-figures in Fig. 5(c)) among four different methods.
\subsection{Ablation Study}
\begin{table}[h!]
	\centering
	\setlength{\tabcolsep}{14pt}
	\caption{Ablation study on texture transformer module.}\label{tab2}
	\begin{tabular}{c|c|c|c}
		\hline
		Method         & HA & SA & PSNR / SSIM     \\ \hline
		Base       &    &    & 33.49 / 0.9131  \\
		SA+Base  &    &  $\checkmark$  & 33.86 / 0.9156  \\
		HA+Base  &  $\checkmark$  &     & 34.07 / 0.9172  \\
		TTM+Base &  $\checkmark$   & $\checkmark$    & 34.25 /  0.9238 \\ \hline
	\end{tabular}
\end{table}
The contribution of the proposed TTM is demonstrated by a set of experiments (i.e. comparison between \textbf{Ref+Original} and \textbf{TTM+Original}) in Table~\ref{tab1} and Fig.~\ref{fig3}. Furthermore, we conduct a detailed ablation study to analyze the effectiveness of different components in the proposed TTM. In this case, we use U-net~\cite{unet} as our \textbf{Base} model. On top of the base model, we progressively add hard attention module (HA) and soft attention module (SA). Ablation results are shown in Table~\ref{tab2}.  As we can see, both HA and SA show positive contributions in term of the reconstruction quality. After adding SA, relevant texture features are enhanced and the less relevant ones are suppressed during feature synthesis. When HA is added, PSNR is improved from 33.49 to 34.07 and SSIM from 0.9131 to 0.9172, which demonstrates the ability of HA to transfer features. It is worth noting that TTM only consists of a few convolutional layers. Its introduced trainable parameters are negligible compared to reconstruction backbones. The detailed network architecture of TTM is provided in the supplementary material.

\section{Conclusion}
We proposed a novel Texture Transformer
Module (TTM) for the MR image reconstruction task which is able to transfer texture features from the reference image to the under-sampled image. The proposed TTM consists of a feature extractor that encourage the joint feature learning for attention computation and two attention modules that can efficiently search and transfer fully-sampled texture feature to under-sampled input image. In addition, the proposed texture transformer module can be directly stacked on the existing MRI reconstruction methods to further improve their performance. Unlike other guidance-based image reconstruction methods, TTM explores the reference images within a single modality and does not require paired auxiliary modalities. Experimental results demonstrate the effectiveness of the proposed TTM among four popular MRI reconstruction approaches on both quantitative and qualitative evaluations. 


\bibliographystyle{IEEEbib}
{\footnotesize{
\bibliography{refs}}}

\end{document}